\documentclass[journal]{IEEEtran}

\usepackage{graphicx}
\usepackage{cite}
\usepackage{picinpar}
\usepackage{amsmath}
\usepackage{url}
\usepackage{flushend}
\usepackage{colortbl}
\usepackage{soul}
\usepackage{multirow}
\usepackage{pifont}
\usepackage{color}
\usepackage{alltt}
\usepackage[hidelinks]{hyperref}
\usepackage{enumerate}
\usepackage{siunitx}
\usepackage{epstopdf}
\usepackage{pbox}
\usepackage{hhline}
\usepackage{amssymb}
\usepackage{stfloats}
\usepackage{float}

\usepackage{xcolor}
\usepackage[ruled,linesnumbered]{algorithm2e}
\usepackage{tabularx}
\usepackage{multirow}
\usepackage{booktabs}
\newcommand{\tabincell}[2]{\begin{tabular}{@{}#1@{}}#2\end{tabular}}

\DeclareSymbolFont{largesymbol}{OMX}{yhex}{m}{n}
\DeclareMathAccent{\Widehat}{\mathord}{largesymbol}{"62}

\newcommand{\red}{\textcolor{red}}

\title{\LARGE \bf
Trajectory Following Strategies for Wireless Capsule Endoscopy under Reciprocally Rotating Magnetic Actuation in a Tubular Environment
}

\author{Yangxin~Xu$^{*}$,~\IEEEmembership{Member,~IEEE,}
        Keyu~Li$^{*}$,~\IEEEmembership{Graduate~Student~Member,~IEEE,}
        Ziqi~Zhao,\\
        and~Max~Q.-H.~Meng$^{\sharp}$,~\IEEEmembership{Fellow,~IEEE}
\thanks{This work is partially supported by National Key R \& D program of China with Grant No. 2019YFB1312400 and Hong Kong RGC CRF grant C4063-18G awarded to Max Q.-H. Meng.}
\thanks{Y. Xu and K. Li are with the Department of Electronic Engineering, the Chinese University of Hong Kong, Hong Kong SAR, China (e-mail: yxxu@link.cuhk.edu.hk; kyli@link.cuhk.edu.hk).}
\thanks{Z. Zhao is with the Department of Electronic and Electrical Engineering, the Southern University of Science and Technology, Shenzhen, China (e-mail: zzq2694@163.com).}
\thanks{Max Q.-H. Meng is with the Department of Electronic and Electrical Engineering of the Southern University of Science and Technology in Shenzhen, China, on leave from the Department of Electronic Engineering, the Chinese University of Hong Kong, Hong Kong SAR, China, and also with the Shenzhen Research Institute of the Chinese University of Hong Kong, Shenzhen, China (e-mail: max.meng@ieee.org).}
\thanks{$^{*}$ The authors contribute equally to this paper.}
\thanks{$^{\sharp}$ Corresponding author.}
}

\begin{document}

\maketitle

\begin{abstract}
Currently used wireless capsule endoscopy (WCE) is limited in terms of inspection time and flexibility since the capsule is passively moved by peristalsis and cannot be accurately positioned.
Different methods have been proposed to facilitate active locomotion of WCE based on simultaneous magnetic actuation and localization technologies.
In this work, we investigate the trajectory following problem of a robotic capsule under rotating magnetic actuation in a tubular environment, in order to realize safe, efficient and accurate inspection of the intestine at given points using wireless capsule endoscopes. Specifically, four trajectory following strategies are developed based on the PD controller, adaptive controller, model predictive controller and robust multi-stage model predictive controller. Moreover, our method takes into account the uncertainty in the intestinal environment by modeling the intestinal peristalsis and friction during the controller design.
We validate our methods in simulation as well as in real-world experiments in various tubular environments, including plastic phantoms with different shapes and an ex-vivo pig colon. The results show that our approach can effectively actuate a reciprocally rotating capsule to follow a desired trajectory in complex tubular environments, thereby having the potential to enable accurate and repeatable inspection of the intestine for high-quality diagnosis.
\end{abstract}

\def\abstractname{Note to Practitioners}
\begin{abstract}
The motivation of this paper is to solve the trajectory following problem for active WCE in the human intestine to realize accurate, efficient and repeatable inspection of the gastrointestinal (GI) tract.
We present four different control strategies for 5-DOF control of a robotic capsule endoscope actuated by a reciprocally rotating permanent magnet to make the capsule follow a predefined trajectory in a tubular environment. The accuracy and efficiency of the approach are validated in simulation and real-world experiments.
The proposed trajectory following strategies can be integrated into existing WCE products to allow automatic and repeatable examination of the GI tract, and can also be extended to the locomotion of other tethered or untethered magnetic devices in the tubular environments for different medical and industrial applications.
In the future, our proposed approach is expected to be combined with image-based automatic diagnosis of the GI tract to provide doctors with better tools for digestive examinations.
\end{abstract}


\section{Introduction}

\IEEEPARstart{A}{ccording} to the data reported in \cite{sung2021global}, cancer is expected to become the leading cause of death in the 21st century and an important barrier to increasing life expectancy worldwide.  Endoscopic screening has been widely adopted to prevent gastrointestinal (GI) cancers by early diagnosis and early treatment \cite{chen2021effectiveness}. However, the commonly used optical endoscopy requires an experienced clinician to manually insert the endoscope and perform inspections, and the procedure usually causes the discomfort of patients \cite{norton2019intelligent}. 
Wireless Capsule Endoscopy (WCE) is a painless and noninvasive solution to visual inspection of the GI tract \cite{iddan2000wireless}, but the whole inspection takes a long time as the capsule is passively moved by peristalsis \cite{meng2004wireless}, and the capsule cannot be accurately positioned in the human body. Therefore, active locomotion and precise localization of a robotic capsule holds great promise to overcome the drawbacks of conventional endoscopy and WCE to enable fast and accurate inspection of the GI tract \cite{ciuti2011capsule}. 
In recent years, simultaneous magnetic actuation and localization (SMAL) have been studied to utilize the magnetic fields to actuate and locate the capsule at the same time. These systems can be divided into coil-based systems and permanent magnet-based systems (e.g., \cite{abbott2020magnetic}. Compared with the coil-based systems (e.g., \cite{yuan2020rectmag3d}), the permanent magnet-based systems are generally more compact, affordable, energy-efficient, and have a larger workspace \cite{pittiglio2019magnetic}, which can provide a easier path toward clinical translation.
The permanent magnet-based systems use an external permanent magnet to actuate a magnetic capsule inside the body, and their magnetic fields are measured by magnetic sensors for capsule localization.

In order to achieve fast and repeatable examination of the intestine with these SMAL systems, it is necessary to develop efficient and accurate methods for trajectory following of the capsule, to track a set of desired waypoints to perform accurate inspection.
In \cite{mahoney2016five}, a PID controller with visual feedback was developed for trajectory following of a capsule in the stomach.
In \cite{taddese2018enhanced}, a trajectory following method was proposed based on a PD controller with the feedback from magnetic sensors and an inertial sensor.
The authors further extended this work to compensate for the capsule's gravity to achieve levitation of the capsule \cite{pittiglio2019magnetic}.
In \cite{barducci2019adaptive}, the capsule's mass was regarded as a slowly varying parameter, and an adaptive controller (AC) was utilized for trajectory following and capsule levitation.
In \cite{scaglioni2019explicit}, the explicit model predictive controller (eMPC) was first used for trajectory following of a capsule.

Different from the aforementioned methods that directly drag the capsule using the magnetic force, some researchers proposed to use a continuously rotating spherical magnet as the actuator for helical propulsion of a magnetic capsule in a tubular environment \cite{mahoney2014generating,popek2017first,popek2016six}.
Other groups have also utilized continuously rotating magnetic actuation in external sensor-based SMAL systems to actuate a capsule in the intestinal environment \cite{xu2020novelsystem,xu2019towards,xu2020novel,xu2020improved}. More recently, a new rotating magnetic actuation method named \textit{reciprocally rotating magnetic actuation} was proposed by Xu et al. \cite{xu2021reciprocally}, to rotate a non-threaded capsule back and forth during propulsion in a narrow tubular environment, in order to reduce the risk of causing intestinal malrotation and enhance patient safety. The reciprocal motion of the capsule has also been demonstrated to help make the intestine stretch open for easier advancement of the capsule.
However, these rotating magnetic actuation based methods have only been developed to propel the capsule through a tubular environment without accurate positioning of the capsule to follow a given trajectory \cite{mahoney2014generating} -- \cite{xu2020improved}.

\begin{figure*}[t]
\setlength{\abovecaptionskip}{-0.0cm}
\centering
\includegraphics[scale=1.0,angle=0,width=0.99\textwidth]{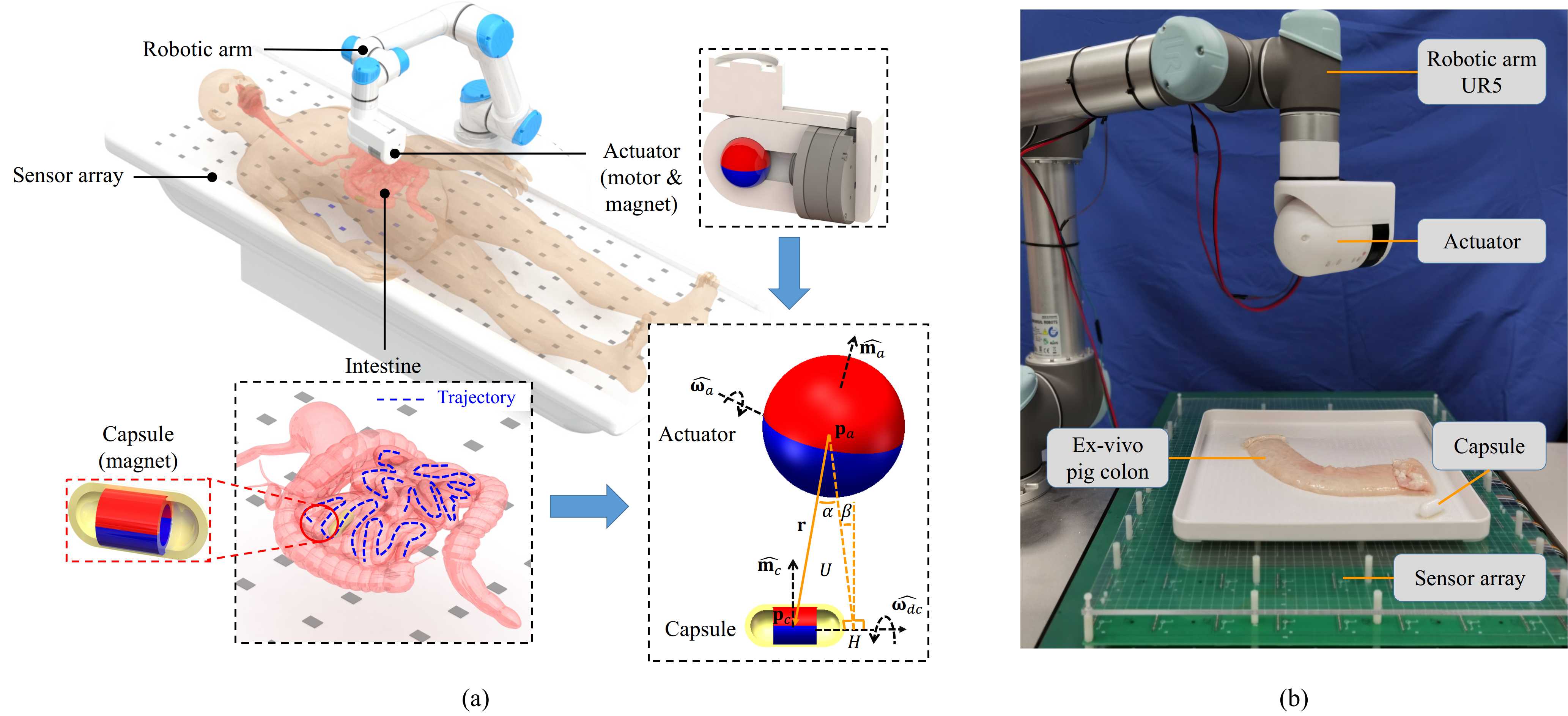}
\caption{(a) The overall design of our SMAL system, which uses an external magnetic sensor array and a reciprocally rotating actuator magnet (mounted at the end-effector of a robotic arm) to realize trajectory following of a mockup magnetic capsule endoscope inside the intestine.  The capsule rotates around $\widehat{\pmb{\omega}_{dc}}$ at $\mathbf{p}_{c}$, and the actuator rotates around $\widehat{\pmb{\omega}_{a}}$ at $\mathbf{p}_{a}$. $\mathbf{p}_{a} H \perp \widehat{\pmb{\omega}_{dc}}$. $\mathbf{r}$ is the vector from $\mathbf{p}_{a}$ to $\mathbf{p}_{c}$. $U$ is the plane formed by $\mathbf{p}_{a}$, $\mathbf{p}_{c}$ and $H$. $\alpha$ is the angle between $\mathbf{r}$ and the $\mathbf{p}_{a} H$, and $\beta$ is the angle between plane $U$ and the vertical line. (b) The real-world system setup in our trajectory following experiments in an ex-vivo pig colon.}
\label{Fig_system3D}
\vspace{-0.3cm}
\end{figure*}

In this work, we are committed to solving the trajectory following problem of a robotic capsule actuated by a rotating magnetic field to safely and efficiently follow a preset trajectory in a narrow tubular environment, in order to realize accurate, efficient and repeatable inspection of the intestine for high-quality diagnosis. An overview of the system is illustrated in Fig. \ref{Fig_system3D}, which is developed based on our previous external sensor-based SMAL systems presented in \cite{xu2020novelsystem}\cite{xu2021adaptive}. 
The mockup capsule endoscope contains a permanent magnetic ring (outer diameter: $12.8mm$, inner diameter: $9mm$, length: $15mm$, N38SH grade) and is actuated by an external spherical permanent magnet (diameter: $50mm$, N42 grade) which is maneuvered by a Universal Robots UR5 robotic arm and a motor. Other functional modules of current WCE (e.g., camera, LEDs, batteries, transmitter and antenna) can be integrated into the magnetic capsule for imaging purposes. The localization is based on an external sensor array composed of $80$ three-axis magnetic sensors (MPU9250, InvenSense), using the localization strategy described in \cite{xu2021adaptive}.

The main contributions of this paper are three-fold:
\begin{itemize}
\item Four control strategies are developed based on two reactive controllers, i.e., PD controller and adaptive controller (AC), and two anticipative controllers, i.e., model predictive controller (MPC) and robust multi-stage model predictive controller (RMMPC), to accomplish the trajectory following task of a robotic capsule endoscope under reciprocal rotating magnetic actuation \cite{xu2021reciprocally} in a tubular environment. 
\item Moreover, we take into consideration the uncertainty in the environment in our controller design by modeling the intestinal peristalsis and friction, in order to better represent the capsule-intestine interaction in the trajectory following task.
\item The proposed trajectory following strategies are evaluated in simulation and real-world experiments on complex-shaped phantoms and an ex-vivo pig colon, and our results have demonstrated the potential of the proposed methods to realize accurate and fast examination of the GI tract using WCE.
\end{itemize}

To our knowledge, this is the first trajectory following method based on rotating magnetic actuation for active WCE in intestinal environments. Also, our method only uses external magnetic sensors to locate the capsule and does not require any magnetic, visual or inertial sensor to be placed in the capsule (e.g., in \cite{taddese2018enhanced}\cite{popek2016six}), which helps reduce the size and power consumption of the capsule and provide a promising path towards clinical translation.

The remainder of this paper is organized as follows. Section II presents the nomenclature and briefly introduces the reciprocally rotating magnetic approach, before the proposed trajectory following strategies are introduced in Section III. Experimental results are presented in Section IV, before we discuss and conclude this work in Section V.

\section{Background}

\subsection{Nomenclature}

Throughout this paper, lowercase normal fonts refer to scalars (e.g., $\mu_{0}$). Lowercase bold fonts refer to vectors (e.g., $\textbf{b}$). The vector with a ``hat'' symbol indicates that the vector is a unit vector of the original vector (e.g., $\widehat{\textbf{r}}$ is the unit vector of $\textbf{r}$), and the vector with an ``[index]'' indicates one component of the vector (e.g., $\textbf{r}[i]$ is the $i$-th component of $\textbf{r}$, $\textbf{r}\in\mathbb{R}^{n\times1}$, $1 \leq i \leq n$). Matrices are represented by uppercase bold fonts (e.g. $\textbf{M}$), and $\textbf{I}_{n}$ denotes $n \times n$ identity matrix. In addition, $Rot_{k}(\theta)$ represents the rotation of $\theta$ degrees around the $+k$-axis, $k\in\{x,y,z\}$.

\subsection{Reciprocally Rotating Magnetic Actuation}

As shown in Fig. \ref{Fig_system3D}(a), the capsule in our system contains a permanent magnetic ring, whose axis coincides with the principal axis of the capsule and is parallel with its dipole moment $\widehat{\mathbf{m}_{c}}$. The capsule is actuated using the reciprocally rotating magnetic actuation (RRMA) approach described in \cite{xu2021reciprocally}, which generates a rotating magnetic field by the extracorporeal actuator magnet to make the capsule reciprocally rotates around its moving direction during the propulsion, in order to reduce the environmental resistance and reduce the risk of causing malrotation of the intestine. Here, we briefly describe the actuation approach in \cite{xu2021reciprocally}. Assume that the capsule rotates around the desired moving direction $\widehat{\pmb{\omega}_{dc}}\in\mathbb{R}^{3 \times 1}$ when the actuator rotates around $\widehat{\pmb{\omega}_{a}}\in\mathbb{R}^{3 \times 1}$. We define the world frame so that $\widehat{\pmb{\omega}_{dc}}$, $\widehat{\pmb{\omega}_{a}}$ are initially aligned with $+x$-axis of the world frame. Then, $\widehat{\pmb{\omega}_{dc}}$ can be represented by $\theta_{cz}$, $\theta_{cy}$ using (\ref{F_wc}):

\begin{equation}
\label{F_wc}
\begin{aligned}
&\quad\ \widehat{\pmb{\omega}_{dc}} = Rot_{z}(\theta_{cz})Rot_{y}(-\theta_{cy})\left[\begin{matrix}1\\0\\0\end{matrix}\right]\\
\Longrightarrow &
\begin{cases}
\begin{split}
\theta_{cz} &= \arctan{\frac{\widehat{\pmb{\omega}_{dc}}[2]}{\widehat{\pmb{\omega}_{dc}}[1]}}, &\theta_{cz} \in [0^{\circ}, 360^{\circ})\quad \\
\theta_{cy} &= \arcsin{\widehat{\pmb{\omega}_{dc}}[3]}, &\theta_{cy} \in (-90^{\circ}, 90^{\circ})
\end{split}
\end{cases}
\end{aligned}
\end{equation}

\noindent and $\widehat{\pmb{\omega}_{a}}$ can be represented by $\theta_{az}$, $\theta_{ay}$ using (\ref{F_wa}):
\begin{equation} 
\label{F_wa}
\begin{aligned}
&\quad\ \widehat{\pmb{\omega}_{a}} = Rot_{z}(\theta_{az})Rot_{y}(-\theta_{ay})\left[\begin{matrix}1\\0\\0\end{matrix}\right]\\
\Longrightarrow &
\begin{cases}
\begin{split}
\theta_{az} &= \arctan{\frac{\widehat{\pmb{\omega}_{a}}[2]}{\widehat{\pmb{\omega}_{a}}[1]}}, &\theta_{az} \in [0^{\circ}, 360^{\circ})\quad \\
\theta_{ay} &= \arcsin{\widehat{\pmb{\omega}_{a}}[3]}, &\theta_{ay} \in (-90^{\circ}, 90^{\circ})
\end{split}
\end{cases}
\end{aligned}
\end{equation}

Assume that the unit magnetic moment of the actuator $\widehat{\mathbf{m}_{a}}$ is initially aligned with $+z$-axis of the world frame, let $\theta_{ax} \in [0^{\circ}, 360^{\circ})$ indicate the angle that $\widehat{\mathbf{m}_{a}}$ rotates around $\widehat{\pmb{\omega}_{a}}$, then $\widehat{\mathbf{m}_{a}}$ can be calculated by (\ref{F_ma}).

\begin{equation}
\label{F_ma}
\widehat{\mathbf{m}_{a}}=Rot_{z}(\theta_{az})Rot_{y}(-\theta_{ay})Rot_{x}(\theta_{ax})\left[\begin{matrix}0\\0\\1\end{matrix}\right]
\end{equation}

In order to generate a rotating magnetic field around the desired rotation axis of the capsule $\widehat{\pmb{\omega}_{dc}}$ at $\mathbf{p}_{c}$, the rotation axis of the actuator $\widehat{\pmb{\omega}_{a}}$ can be calculated by (\ref{F_rotating_actuation}) \cite{mahoney2014generating}.

\begin{equation}
\label{F_rotating_actuation}
\widehat{\pmb{\omega}_{a}}=\Widehat{\left((3\widehat{\mathbf{r}}{\widehat{\mathbf{r}}}^{T}-\mathbf{I}_{3})\ \widehat{\pmb{\omega}_{dc}}\right)}
\end{equation}

\noindent where $\mathbf{r}=\mathbf{p}_{c}-\mathbf{p}_{a}$ is the position of the capsule relative to the actuator, then $\mathbf{r}$ can be represented by 


\begin{equation}
\label{F_r}
\mathbf{r}=d\left(Rot_{z}(\theta_{cz})Rot_{y}(-\theta_{cy})Rot_{x}(\beta)Rot_{y}(\alpha)\left[\begin{matrix}0\\0\\-1\end{matrix}\right]\right)
\end{equation}

\noindent where $d=\|\mathbf{r}\|=\|\mathbf{p}_{c}-\mathbf{p}_{a}\|$ is the distance between the capsule and the actuator. $\alpha$, $\beta$ are illustrated in Fig. \ref{Fig_system3D}(a).


The RRMA model \cite{xu2021reciprocally} shows that when the actuator reciprocally rotates a relatively small angle $\theta_{ar}$ around $\theta_{ax} = 180^{\circ}$, the magnetic force applied to the capsule $\mathbf{f}$ can be approximated using (\ref{F_f_theta_c_approximate}):

\begin{equation}
\label{F_f_theta_c_approximate}
\begin{split}
\mathbf{f}(d,\alpha,\beta,\theta_{ax},\ &\widehat{\pmb{\omega}_{dc}}) \doteq \mathbf{f}(d,\alpha,\beta,\theta_{ax}=180^{\circ},\widehat{\pmb{\omega}_{dc}}),\\
\theta_{ax} &\in [180^{\circ}-\theta_{ar},180^{\circ}+\theta_{ar}]
\end{split}
\end{equation}

\section{Methods}

\subsection{Overview}

An overview of the proposed trajectory following algorithm with $4$ control strategies can be found in Algorithm \ref{Alg_trajectoryfollowing}. In the remainder of this section, we first describe the 5D control of a capsule under RRMA, and then introduce our method to model the environmental resistance in the intestinal environment, before presenting the reactive-controller-based and model-predictive-controller-based trajectory following strategies.

\subsection{5D Control of the Capsule under RRMA}

Similar to \cite{mahoney2016five}, our system can realize 5D control of the capsule, including 3D control in force and 2D control in orientation (the capsule cannot be controlled to rotate a specific angle around its principal axis as the capsule is always reciprocally rotating around this axis under RRMA.) As discussed before, the current rotation axis of the capsule $\widehat{\pmb{\omega}_{c}}$ in the narrow tubular luman may not always be aligned with the desired rotation axis $\widehat{\pmb{\omega}_{dc}}$. If the angle between $\widehat{\pmb{\omega}_{c}}$ and $\widehat{\pmb{\omega}_{dc}}$ is too large ($\Phi>\Phi_{th}$), the capsule will not rotate normally under magnetic actuation. To deal with this problem, we propose a Spherical Linear Interpolation (SLI) based method to linearly generate the next heading direction $\widehat{\pmb{\omega}_{nc}}$ with respect to angle $\Phi$ using (\ref{F_wnc}):

\begin{equation}
\label{F_wnc}
\begin{split}
\widehat{\pmb{\omega}_{nc}}=
\begin{cases}
\widehat{\pmb{\omega}_{dc}}&,\Phi \leq \Phi_{th}\\
\frac{\sin(\Phi-\Phi_{th})}{\sin(\Phi)}\widehat{\pmb{\omega}_{c}}+\frac{\sin(\Phi_{th})}{\sin(\Phi)}\widehat{\pmb{\omega}_{dc}}&,\Phi>\Phi_{th}
\end{cases}
\end{split}
\end{equation}

\noindent where $\Phi$ is the angle between $\widehat{\pmb{\omega}_{c}}$ and $\widehat{\pmb{\omega}_{dc}}$, and the threshold $\Phi_{th}$ is set to $45^{\circ}$. When $\Phi \leq \Phi_{th}$, the desired heading direction $\widehat{\pmb{\omega}_{dc}}$ is considered achievable and directly used as $\widehat{\pmb{\omega}_{nc}}$. When $\Phi>\Phi_{th}$, $\frac{\sin(\Phi-\Phi_{th})}{\sin(\Phi)}$ and $\frac{\sin(\Phi_{th})}{\sin(\Phi)}$ are used as the weights to combine $\widehat{\pmb{\omega}_{c}}$ and $\widehat{\pmb{\omega}_{dc}}$ to calculate $\widehat{\pmb{\omega}_{nc}}$.

According to (\ref{F_f_theta_c_approximate}), given the desired magnetic force $\mathbf{f}_{d}$ and the next heading direction $\widehat{\pmb{\omega}_{nc}}$ as the modified goal, the configuration of the actuator ($d$, $\alpha$, $\beta$) can be determined by solving the optimization problem in (\ref{F_f_config}). We apply the Trust Region Reflective algorithm \cite{branch1999subspace} to solve this problem.

\begin{equation}
\label{F_f_config}
\begin{aligned}
\mathop{\arg\min}_{d,\alpha,\beta} \quad & \|\mathbf{f}_{d}-\mathbf{f}(d,\alpha,\beta,\theta_{ax}=180^{\circ},\widehat{\pmb{\omega}_{nc}})\|\\
\textrm{subject to} \quad & d\in[0.10m,0.25m],\\
& \alpha,\beta\in[-15^{\circ},15^{\circ}]\\
\end{aligned}
\end{equation}

\subsection{Modeling of Environmental Resistance in the Intestine} \label{section_MMC}

In the complex intestinal environment, the environmental resistance has a non-negligible impact on the movement of the capsule, which should be properly modeled for the design of trajectory following algorithms. 
In this work, we assume that the magnitude of friction exerted on the capsule by a static intestine is:


\begin{equation}
\label{F_friction}
\mathbf{f}_{fric} = -\rho_{fric} \frac{\dot{\mathbf{p}}_{c}}{\|\dot{\mathbf{p}}_{c}\|}, \\
\end{equation}

The real human intestine is a complex dynamic environment due to the physiological motions such as peristalsis, which occurs as a series of wave-like contraction and relaxation of muscles that move food through from the stomach to the distal end of the colon. The electromechanical activity that triggers peristaltic waves is called the Migrating motor complex (MMC) \cite{takahashi2013interdigestive}. MMC periodically occurs every $90 \sim 120$ minutes in humans and each cycle can be divided into four phases \cite{takahashi2013interdigestive}. Phase~I is a quiescent period with rare contractions, which lasts about half of the cycle. Phase~II lasts about $1/4$ of the cycle, during which intermittent low-amplitude contractions will occur. Phase~III is a period of about $5 \sim 10$ minutes that consists of short burst of strong, regular contractions. Phase~IV is a short transition period between Phase~III and Phase~I to repeat the cycle. Stronger contractions will result in greater pressure on the capsule, thereby increasing the friction. Therefore, we modify (\ref{F_friction}) by introducing a coefficient $R$ to model the impact of the MMC (i.e., for Phase I, $R=1.0$; for Phase III, $R=R_{max}$ since high-amplitude contractions occur; for Phase II and Phase IV, $R=\frac{1.0+R_{max}}{2}$). In addition, a random disturbance $\mathbf{f}_{dist}$ with an upper boundary $\rho_{dist}$ is added to represent the uncertainty of the environment. Finally, the environmental resistance modeled in our work become


\begin{equation}
\label{F_env}
\begin{aligned}
&\mathbf{f}_{env}=R\ \mathbf{f}_{fric} + \mathbf{f}_{dist},\\
\text{where }&R=\begin{cases}
\begin{split}
&\ \ \ \ \ \ 1.0 &, \text{Phase I} \\
&(1.0+R_{max})/2  &, \text{Phase II or IV}\\
&\ \ \ \ \ R_{max} &, \text{Phase III}
\end{split}
\end{cases}, \\
&\mathbf{f}_{dist} \in B(\mathbf{0},\rho_{dist})
\end{aligned}
\end{equation}


\noindent where $B(\mathbf{0},\rho_{dist})=\{x \in \mathbb{R}^{3}, \|x\|_{2}\leq \rho_{dist}\}$ is an Euclidean ball in $\mathbb{R}^{3}$.

\begin{figure*}[t]
\centering
\includegraphics[scale=1.0,angle=0,width=0.99\textwidth]{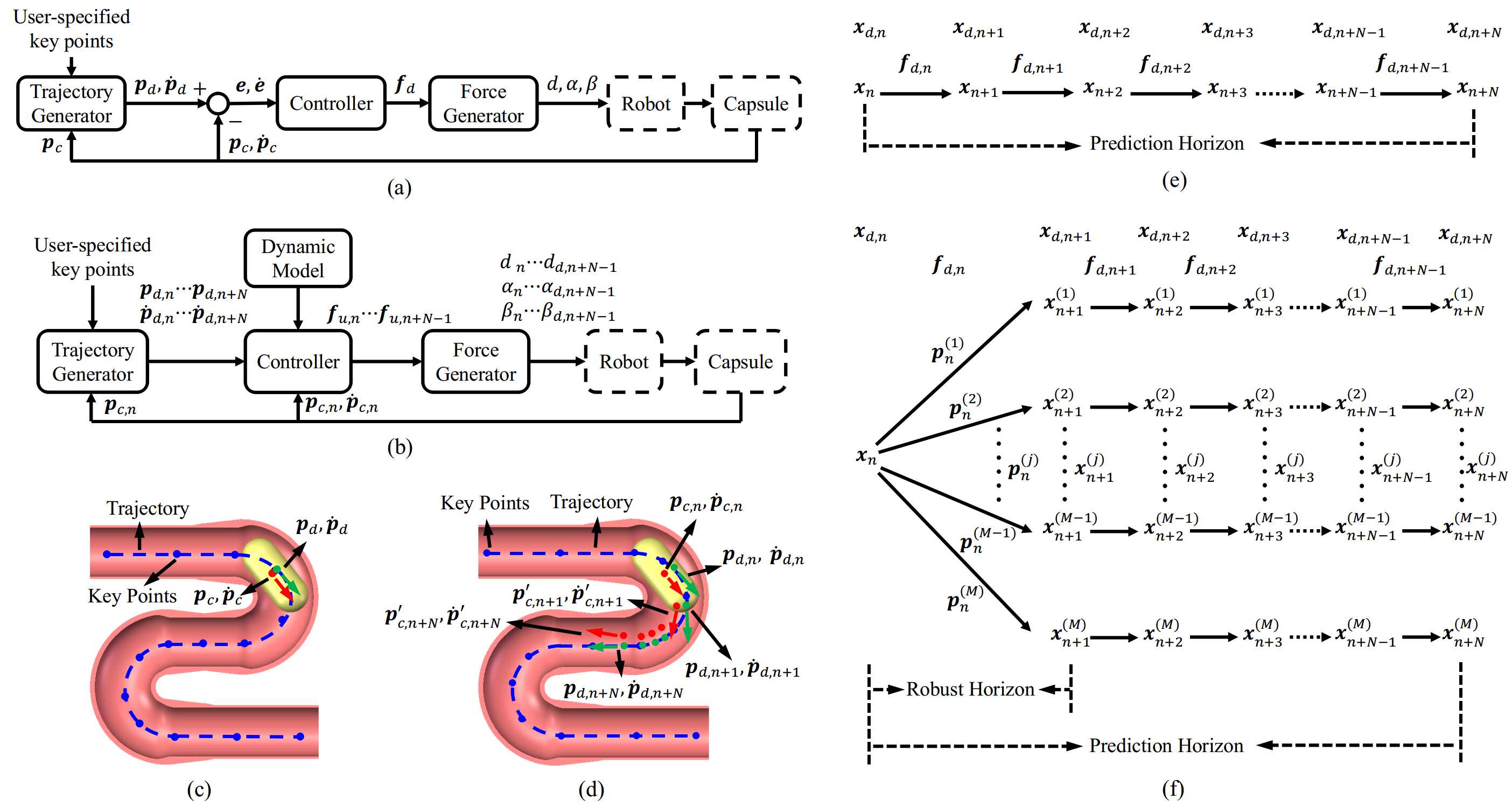}
\caption{(a) and (b) show the overall scheme of the (a) reactive controller- and (b) model predictive controller (MPC) based trajectory following approaches. In (c)(d), the desired trajectory (blue dashed line) is generated by interpolation through user-specified key points (blue points). In (c), $\mathbf{p}_{d}$,$\dot{\mathbf{p}}_{d}$ are the desired position and velocity of the capsule, and $\mathbf{p}_{c}$ is the current position of the capsule. In (d), the green arrows represent the desired position and velocity sequences ($\mathbf{p}_{d,n} \cdots \mathbf{p}_{d,n+N}$,$\dot{\mathbf{p}}_{d,n}$), and the red arrows illustrate the current position ($\mathbf{p}_{c,n}$) and the estimated future positions ($\mathbf{p}^{\prime}_{c,n+1} \cdots \mathbf{p}^{\prime}_{c,n+N}$). (e) shows the state transition in MPC with a prediction horizon of $N$. (b) shows the state transitions in the Robust Multi-stage Model Predictive Controller (RMMPC), where the prediction horizon is $N$ and the robust horizon is $1$.}
\label{Fig_trajectoryfollowing}
\end{figure*}

\begin{algorithm}[t] \small 
\caption{Trajectory Following Algorithm}
\label{Alg_trajectoryfollowing}
\KwIn{current position, velocity, moving direction of the capsule ($\mathbf{p}_{c}$, $\dot{\mathbf{p}}_{c}$, $\widehat{\pmb{\omega}_{c}}$), and \textit{controller}}
\KwOut{configuration of the actuator ($d$, $\alpha$, $\beta$)}
\Switch{controller}{
\Case{"PD" or "AC"}
{Calculate the desired position and velocity of the capsule $\mathbf{p}_{d}$, $\dot{\mathbf{p}}_{d}$ by (\ref{F_traj_generator})(\ref{F_traj_generator_dot})\;
\Switch{controller}{
\Case{"PD"}{Calculate  desired magnetic force $\mathbf{f}_{d}$ by (\ref{F_PDC})\;}
\Case{"AC"}{Calculate  desired magnetic force $\mathbf{f}_{d}$ by (\ref{F_AC})\;}
}
Obtain the desired moving direction $\widehat{\pmb{\omega}_{dc}}=\frac{\dot{\mathbf{p}_{d}}}{\|\dot{\mathbf{p}_{d}}\|}$\;
}
\Case{"MPC" or "RMMPC"}
{Calculate  desired position and velocity sequences of the capsule $\mathbf{p}_{d,0}\cdots\mathbf{p}_{d,N}$, $\dot{\mathbf{p}}_{d,0}\cdots\dot{\mathbf{p}}_{d,N}$ by (\ref{F_traj_generator_MPC})\;
\Switch{controller}{
\Case{"MPC"}{Calculate  desired magnetic force sequence $\mathbf{f}_{d,0}\cdots\mathbf{f}_{d,N-1}$ by (\ref{MPC_V})\;}
\Case{"RMMPC"}{Calculate  desired magnetic force sequence $\mathbf{f}_{d,0}\cdots\mathbf{f}_{d,N-1}$ by (\ref{RMPC_V})\;}
}
Obtain the desired moving direction $\widehat{\pmb{\omega}_{dc}}=\frac{\dot{\mathbf{p}}_{d,0}}{\|\dot{\mathbf{p}}_{d,0}\|}$\;
Obtain the desired magnetic force $\mathbf{f}_{d}=\mathbf{f}_{d,0}$\;
}
}
Calculate the next moving direction $\widehat{\pmb{\omega}_{nc}}$ by (\ref{F_wnc})\;
Determine the configuration of the actuator ($d$,$\alpha$,$\beta$) by (\ref{F_f_config})\;
\Return{configuration of the actuator} $d$, $\alpha$, $\beta$\;
\end{algorithm}

\subsection{Reactive Controller based Trajectory Following Strategies}

We first develop a closed-loop trajectory following algorithm based on two reactive controllers, i.e., PD Controller and Adaptive Controller (AC), as shown in Fig. \ref{Fig_trajectoryfollowing}(a). As shown in Fig. \ref{Fig_trajectoryfollowing}(c), after the user specifies some key points to describe the desired trajectory, a smooth trajectory is generated through the points with cubic spline interpolation \cite{dierckx1995curve}. During the trajectory following step, given the current position of the capsule $\mathbf{p}_{c}$, the next desired position of the capsule $\mathbf{p}_{d}$ can be obtained as the nearest point to $\mathbf{p}_{c}$ on the trajectory by solving the optimization problem in (\ref{F_traj_generator}).

\begin{equation}
\label{F_traj_generator}
\begin{aligned}
\mathop{\arg\min}_{s} \quad & \|\mathbf{p}_{d} - \mathbf{p}_{c}\|\\
\textrm{subject to} \quad & \mathbf{p}_{d} = \mathbf{p}_{traj}(s),\\
& s \in [0.0,1.0]\\
\end{aligned}
\end{equation}
\vspace{0.2cm}

\noindent where $\mathbf{p}_{traj}(s)$, $s \in [0.0,1.0]$ denotes any point on the desired trajectory. Let $\dot{\mathbf{p}}_{traj}(s)$ denote the first-order derivative of $\mathbf{p}_{traj}(s)$, then the desired orientation of the capsule is

\begin{equation}
\label{F_traj_generator_dot1}
\widehat{\pmb{\omega}_{dc}}=\frac{\dot{\mathbf{p}}_{traj}(s)}{\|\dot{\mathbf{p}}_{traj}(s)\|}
\end{equation}
\vspace{0.2cm}

\noindent which is then modified as $\widehat{\pmb{\omega}_{nc}}$ using (\ref{F_wnc}). Let $V_{c}$ be the pre-set constant speed of the capsule, then the next desired velocity of the capsule $\dot{\mathbf{p}}_{d}$ can be determined with (\ref{F_traj_generator_dot}):


\begin{equation}
\label{F_traj_generator_dot}
\dot{\mathbf{p}}_{d}=V_{c}\ \widehat{\pmb{\omega}_{nc}}\\
\end{equation}

The force experienced by the capsule is composed of the magnetic force $\mathbf{f}_{d}$, the capsule's gravity $\mathbf{f}_{g}$, and the environmental resistance $\mathbf{f}_{env}$. Therefore, the dynamics of the capsule is presented as follows:

\begin{equation}
\label{F_dynamics}
\begin{aligned}
m_{c}\ddot{\mathbf{p}}_{c} = \mathbf{f}_{d} + \mathbf{f}_{g} + \mathbf{f}_{env} = \mathbf{f}_{d} + \mathbf{f}_{g} + R\ \mathbf{f}_{fric} + \mathbf{f}_{dist}\\
\end{aligned}
\end{equation}
\vspace{0.2cm}

Let $\mathbf{e}=\mathbf{p}_{d}-\mathbf{p}_{c}$ and $\dot{\mathbf{e}}=\dot{\mathbf{p}_{d}}-\dot{\mathbf{p}_{c}}$ refer to the errors in position and velocity, respectively. The two reactive controllers for the trajectory following of the capsule are designed as follows:

\subsubsection{PD controller}

In the design of the PD controller, we assume that the coefficient $R$ is a constant ($R=1.0$) and the disturbance $\mathbf{f}_{dist}$ can be neglected, so that the proposed PD controller in (\ref{F_PDC}) can make the system stable.

\begin{equation}
\label{F_PDC}
\mathbf{f}_{d} = \mathbf{K}_{P}\mathbf{e} + \mathbf{K}_{D}\dot{\mathbf{e}} - \mathbf{f}_{g} - \mathbf{f}_{fric},
\end{equation}
\vspace{0.4cm}
where $\mathbf{K}_{P}$ and $\mathbf{K}_{D}$ are positive definite matrices.

\subsubsection{Adaptive controller (AC)}

It is assumed that the the coefficient $R$ is slowly varying ($\dot{R} \approx 0$) and the disturbance $\mathbf{f}_{dist}$ is also neglected, so that the proposed AC in (\ref{F_AC}) can make the system stable.
\begin{equation}
\label{F_AC}
\mathbf{f}_{d} = \mathbf{K}_{P}\mathbf{e} + \mathbf{K}_{D}\dot{\mathbf{e}} - \mathbf{f}_{g} + \left(\int_{0}^{t}\dot{\mathbf{e}}^{T}\mathbf{f}_{fric}\right)\mathbf{f}_{fric},
\end{equation}
\vspace{0.3cm}
where $\mathbf{K}_{P}$ and $\mathbf{K}_{D}$ are positive definite matrices.

After obtaining the desired force $\mathbf{f}_{d}$, the actuator pose can be determined by solving ($d$, $\alpha$, $\beta$) using (\ref{F_f_config}). 
The proof of system stability using these controllers can be easily obtained using Lyapunov's second method for stability \cite{lyapunov1992general}.

\subsection{Model Predictive Controller based Trajectory Following Strategies}

In addition to reacting to current errors, the model predictive controllers (MPCs) can also anticipate the future behavior of the system over finite time window (prediction horizon) by using the dynamic model of the system. The MPCs are also advantageous in that they can explicitly consider the hard and soft constraints of outputs and states.

Fig. \ref{Fig_trajectoryfollowing}(b) illustrates our proposed workflow for closed-loop trajectory following of the capsule based on the MPCs. Let $f_{c}$ denote the control frequency. The controller predicts the future states within a time interval of $\frac{N}{f_{c}}$, where $N$ is the prediction horizon. 
Let $\mathbf{p}_{c,n}$, $\mathbf{p}_{d,n}$  be the current and desired positions of the capsule and $\dot{\mathbf{p}}_{c,n}$, $\dot{\mathbf{p}}_{d,n}$ be the velocities, the current state of the system is defined as $\mathbf{x}_{n}=\left[\mathbf{p}_{c,n}, \dot{\mathbf{p}}_{c,n}\right]^T$, and the desired future states over the prediction horizon $N$ are ($\mathbf{x}_{d,n}$, $\cdots$, $\mathbf{x}_{d,n+N}$), where $\mathbf{x}_{d,n+i}=\left[\mathbf{p}_{d,n+i}, \dot{\mathbf{p}}_{d,n+i}\right]^T$. As shown in Fig. \ref{Fig_trajectoryfollowing}(d), given the current position of the capsule $\mathbf{p}_{c,n}$, the $N+1$ points $\mathbf{p}_{d,n}\cdots\mathbf{p}_{d,n+N}$ and their first-order derivatives $\dot{\mathbf{p}}_{d,n}\cdots\dot{\mathbf{p}}_{d,n+N}$ can be found on the generated trajectory using an optimization based searching algorithm in (\ref{F_traj_generator_MPC}).

\begin{align}
\mathop{\arg\min}_{s_{n+i},i \in \{0,\cdots,N\}} \quad & \|\mathbf{p}_{d,n+i} - \mathbf{p}_{c,n+i}^{\prime}\| \label{F_traj_generator_MPC} \\
\textrm{subject to} \quad
& \mathbf{p}_{c,n}^{\prime} = \mathbf{p}_{c,n}, \notag \\
& \mathbf{p}_{d,n+i} = \mathbf{p}_{traj}(s_{n+i}), \notag \\
& s_{n+i} \in [0.0,1.0], \notag \\
&\dot{\mathbf{p}}_{d,n+i}=V_{c}\ \frac{\dot{\mathbf{p}}_{traj}(s_{n+i})}{\|\dot{\mathbf{p}}_{traj}(s_{n+i})\|}, \notag \\
& \mathbf{p}_{c,n+i+1}^{\prime} = \mathbf{p}_{d,n+i}+\dot{\mathbf{p}}_{d,n+i}\ \frac{1}{f_{c}}, i < N \notag
\end{align}

\noindent where $\mathbf{p}_{traj}(s_{n+i})$ and $\dot{\mathbf{p}}_{traj}(s_{n+i})$ denote any point and its first derivative on the generated trajectory, respectively. $V_{c}$ is the pre-set constant speed of the capsule. $\mathbf{p}_{c,n+i+1}^{\prime}$ is used to roughly estimate the next position of the capsule.

\subsubsection{Model predictive controller (MPC)} \label{MPC-based}

As shown in Fig. \ref{Fig_trajectoryfollowing}(e), the state evolution in MPC is assumed to have no uncertainty, i.e., the dynamic model is simplified to set $R=1.0$ and neglect the disturbance. Given the current state $\mathbf{x}_{n}$ and the desired states ($\mathbf{x}_{d,n}$, $\cdots$, $\mathbf{x}_{d,n+N}$) over the prediction horizon $N$, the desired forces ($\mathbf{f}_{d,n}$, $\cdots$, $\mathbf{f}_{d,n+N-1}$) applied to the capsule during $N$ time steps can be obtained by solving the optimization problem in (\ref{MPC_V}) to minimize the errors between the estimated sequence of future states ($\mathbf{x}_{n}, \cdots, \mathbf{x}_{n+N}$) and the desired states ($\mathbf{x}_{d,n}, \cdots, \mathbf{x}_{d,n+N}$).

\begin{align}
\mathop{\arg\min}_{\mathbf{f}_{d,n}\cdots\mathbf{f}_{d,n+N-1}} \quad & {\mathbf{e}^{T}_{n+N}}\mathbf{W}_{N}\mathbf{e}_{n+N} +\sum_{i=0}^{N-1}\left({\mathbf{e}^{T}_{n+i}}\mathbf{W}_{x}\mathbf{e}_{n+i}\right.  \notag \\
& \left. +\Delta\mathbf{f}_{n+i}^{T}\mathbf{W}_{f}\Delta\mathbf{f}_{n+i} \right) \label{MPC_V}  \\
\textrm{subject to} \quad
&\mathbf{e}_{n+N} = \mathbf{x}_{d,n+N}-\mathbf{x}_{n+N}, \label{RMPC_estatefin}\\
&\mathbf{e}_{n+i} = \mathbf{x}_{d,n+i}-\mathbf{x}_{n+i}, \label{RMPC_estate}\\
&\Delta\mathbf{f}_{n+i} = \mathbf{f}_{d,n+i+1}-\mathbf{f}_{d,n+i}, \label{RMPC_eforce}\\
&\mathbf{x}_{n+i+1} = \Phi \left(\mathbf{x}_{n+i},\mathbf{f}_{d,n+i},R\right), \label{RMPC_model}\\
& \mathbf{x}_{min} \leq \mathbf{x}_{n+i} \leq \mathbf{x}_{max}, \label{RMPC_statelim}\\
& f_{min} \leq \| \mathbf{f}_{d,n+i} \| \leq f_{max}, \label{RMPC_forcelim}\\
& i \in \{0,1,\cdots,N-1\} \label{RMPC_nlim}
\end{align}
\vspace{0.2cm}

The objective in (\ref{MPC_V}) contains three terms. The first term tries to minimize the final state error $\mathbf{e}_{n+N}$, the second term tries to minimize the middle-state errors $\mathbf{e}_{n+i}$, and the last term tries to minimize the change between successive output forces $\Delta\mathbf{f}_{n+i}$. $\mathbf{W}_{N}$, $\mathbf{W}_{x}$, and $\mathbf{W}_{f}$ are the weights used to combine the three terms.
As shown in (\ref{RMPC_model}), each future state $\mathbf{x}_{n+i+1}$ within the prediction horizon $N$ can be estimated with a function $\Phi$ based on the dynamic model of the system, given the previous state $\mathbf{x}_{n+i}$, the desired output force $\mathbf{f}_{d,n+i}$ and the MMC-related coefficient $R$. At time step $n+i$, $\Phi$ can be represented by (\ref{F_dynamics_MPC}):

\begin{equation}
\label{F_dynamics_MPC}
\begin{aligned}
\dot{\mathbf{p}}_{c,n+i+1} &= \dot{\mathbf{p}}_{c,n+i} + \ddot{\mathbf{p}}_{c,n+i} \frac{1}{f_{c}}\\
\mathbf{p}_{c,n+i+1} &= \mathbf{p}_{c,n+i} + \dot{\mathbf{p}}_{c,n+i} \frac{1}{f_{c}} + \frac{1}{2} \ddot{\mathbf{p}}_{c,n+i} \frac{1}{f_{c}^{2}}\\
m_{c}\ddot{\mathbf{p}}_{c,n+i} &= \mathbf{f}_{d,n+i} + \mathbf{f}_{g} + R\ \mathbf{f}_{fric} + \mathbf{f}_{dist}\\
\end{aligned}
\end{equation}
\vspace{0.2cm}

Besides, (\ref{RMPC_statelim}) and (\ref{RMPC_forcelim}) are used to indicate the position limit of the capsule in the defined workspace and the force limit for safety reasons \cite{zhang2020experimental}.

\subsubsection{Multi-stage model predictive controller (RMMPC)} \label{section_RMMPC}
In order to take into consideration the uncertainty of the intestinal environment on the capsule movement (as discussed in Section~\ref{section_MMC}), we develop an advanced version of the MPC-based trajectory following approach based on the robust multi-stage model predictive controller (RMMPC) \cite{lucia2013multi}.

As shown in Fig. \ref{Fig_trajectoryfollowing}(f), the state transitions of RMMPC can be represented as a tree, where each path from the root node to one leaf node represents one possible scenario defined by one possible realization of uncertain parameters. The $j$-th possible realization of uncertain parameters at time step $n$ is defined as $\mathbf{p}^{(j)}_{n}$, the initial state at time step $n$ is $\mathbf{x}_{n}$, and the future states over the prediction horizon of $N$ under the $j$-th possible realization are $\mathbf{x}^{(j)}_{n+i}$, $i \in \{1,\cdots,N\}$. The number of steps where branching is applied is called the robust horizon $N_{robust}$, and the uncertain parameters are fixed for the last $N-N_{robust}$ steps to maintain the computational tractability. Here we set $N_{robust}$ to 1 to take into consideration the $4$ possible realizations of the MMC-related parameter $R$ (corresponding to the four phases of MMC) at the first time step. Given the current state $\mathbf{x}_{n}$ and the desired states ($\mathbf{x}_{d,n}$, $\cdots$, $\mathbf{x}_{d,n+N}$), the optimal required forces ($\mathbf{f}_{d,n}$, $\cdots$, $\mathbf{f}_{d,n+N-1}$) can be obtained by solving the optimization problem in (\ref{RMPC_V}).

\begin{align}
\mathop{\arg\min}_{\mathbf{f}_{d,n}\cdots\mathbf{f}_{d,n+N-1}} \quad & \sum_{j=1}^{M}\  w_{j} \left\{ {\mathbf{e}_{n+N}^{(j)}}^{T}\mathbf{W}_{N}\mathbf{e}_{n+N}^{(j)} \right.  \notag \\
& +\sum_{i=0}^{N-1}\left({\mathbf{e}_{n+i}^{(j)}}^{T}\mathbf{W}_{x}\mathbf{e}_{n+i}^{(j)}\right.  \notag \\
& \left.\left. +\Delta\mathbf{f}_{n+i}^{T}\mathbf{W}_{f}\Delta\mathbf{f}_{n+i} \right) \right\} \label{RMPC_V}  \\
\textrm{subject to} \quad
&\mathbf{e}_{n+N}^{(j)} = \mathbf{x}_{d,n+N}^{(j)}-\mathbf{x}_{n+N}^{(j)}, \notag\\
&\mathbf{e}_{n+i}^{(j)} = \mathbf{x}_{d,n+i}^{(j)}-\mathbf{x}_{n+i}^{(j)}, \notag\\
&\Delta\mathbf{f}_{n+i} = \mathbf{f}_{d,n+i+1}-\mathbf{f}_{d,n+i}, \notag\\
&\mathbf{x}^{(j)}_{n+i+1} = \Phi \left(\mathbf{x}^{(j)}_{n+i},\mathbf{f}_{d,n+i},R^{(j)}\right), \notag\\
& \mathbf{x}_{min} \leq \mathbf{x}_{n+i}^{(j)} \leq \mathbf{x}_{max}, \notag\\
& f_{min} \leq \| \mathbf{f}_{d,n+i} \| \leq f_{max}, \notag\\
& i \in \{0,1,\cdots,N-1\}, \notag\\
& j \in \{1,2,\cdots,M\} \notag
\end{align}
\vspace{0.1cm}

\noindent where $j \in \{1,2,\cdots,M\}$ indicates that the variable corresponds to the $j$-th possible realization of uncertain parameters, and  $w_{j}$ in the objective function represents the weight to combine each possible scenario. The definition of other constraints in (\ref{RMPC_V}) are the same as that in (\ref{MPC_V}). Different from the MPC-based approach, we do not assume $R$ to be a constant, but use the method introduced in Section~\ref{section_MMC} and (\ref{F_env}) to determine the value of $R^{(j)}$ in the RMMPC-based approach to take into account the impact of MMC-triggered peristalsis on the capsule. For simplicity, we use probability to roughly estimate the current MMC phase to determine the value of $R$ (i.e., the probabilities for Phase I and Phase III to occur are 50\% and 5\%, respectively, and Phase II and IV have a total chance of 45\% to occur.)

\section{Experiments and Results}

\subsection{Simulation Experiments}

\begin{figure*}[t]
\centering
\includegraphics[scale=1.0,angle=0,width=0.99\textwidth]{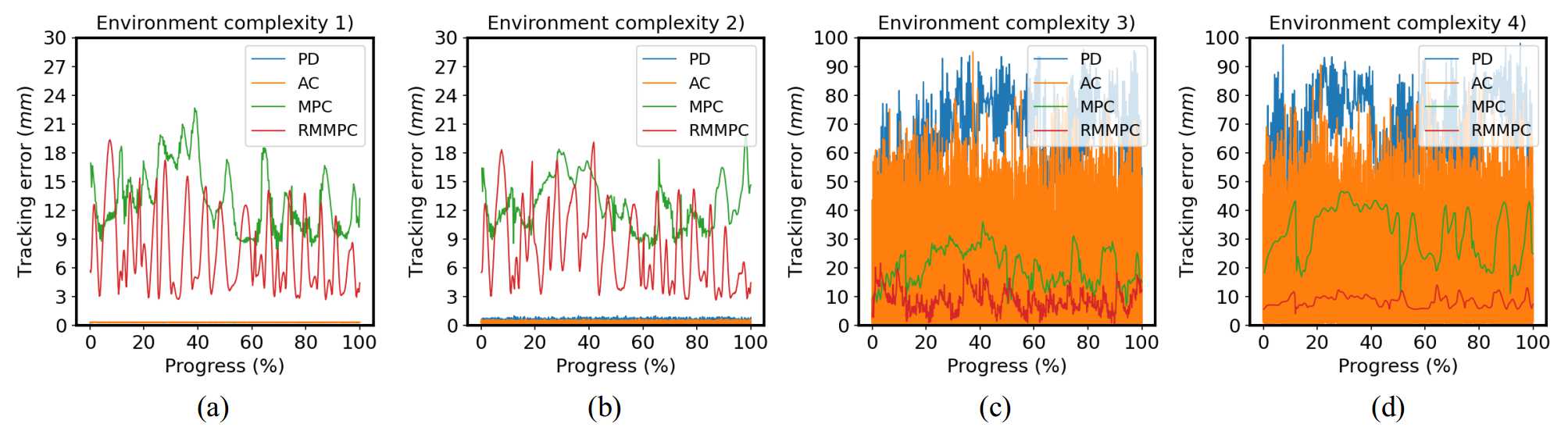}
\caption{Tracking errors of the four control strategies during the trajectory following in the simulated intestinal environments with increasing complexities.}
\label{Fig_simulation}
\end{figure*}

We first performed a preliminary test of the proposed trajectory following strategies in a simulation environment built with Python to simulate the magnetic locomotion of a capsule during the trajectory following task in a simulated intestine. The desired trajectory is set based on the shape of the real human small intestine \cite{elaine2017essentials} with a reduced length of $2.46m$. The environmental resistance in the simulated intestine is modeled in $4$ different settings with increased complexity: 

\begin{enumerate}
\item  The \textit{ideal} situation, where the MMC-related coefficient $R\equiv1.0$ and the environmental disturbance $\mathbf{f}_{dist} \equiv \mathbf{0}$; 
\item Environment with slowly varying $R$ ($1.0 \leq R \leq 2.0$, $\dot{R} \approx 0$) and $\mathbf{f}_{dist} \equiv \mathbf{0}$; 
\item Environment with varying $R$ caused by the MMC mechanism (modeled with the probability-based method introduced in Section~\ref{section_RMMPC}), and $\mathbf{f}_{dist} \equiv \mathbf{0}$; and 
\item Environment with varying $R$ caused by the MMC mechanism, and varying $\mathbf{f}_{dist}$ with a maximum magnitude of $5mN$. In all settings, $\rho_{fric}$ in (\ref{F_friction}) is set to $50mN$ and the speed of the capsule is set to $3mm/s$. 5 trials were conducted in each environment.
\end{enumerate}

\begin{table}[tb] \renewcommand\arraystretch{1.2}
\centering
\caption{Mean Tracking Accuracy in the Simulated Intestinal Environments with Increasing Complexities}
\begin{tabular}{|m{1.5cm}<{\centering} |m{1.2cm}<{\centering} |m{1.2cm}<{\centering} |m{1.2cm}<{\centering} |m{1.2cm}<{\centering}| }
\hline
\multirow{2}{*}{\tabincell{c}{\textbf{Controller}}} & \multicolumn{4}{c|}{\textbf{Simulation environment}} \\
\cline{2-5}
{} &  \textbf{1)} &  \textbf{2)} &  \textbf{3)} & \textbf{ 4)} \\
\hline
\hline
{PD} &  $0.3mm$  & $0.5mm$  & $64.9mm$  & $66.5mm$ \\
\hline
{AC} &  $0.3mm$   & $0.3mm$    & $11.9mm$  & $13.9mm$ \\
\hline
{MPC}  & $13.1mm$  & $12.6mm$   & $20.1mm$  & $32.0mm$ \\
\hline
{RMMPC}  & $7.7mm$ & $8.1mm$  & $8.5mm$  & $8.3mm$ \\
\hline
\end{tabular}
\label{T_simulation}
\end{table}

The tracking accuracy in the simulated intestinal environments with different complexity settings when using the four presented controllers are shown in Fig. \ref{Fig_simulation} and Table \ref{T_simulation}. It can be observed that in the static and slowly changing environments 1) and 2), the reactive controllers (PD and AC) can achieve better tracking accuracy compared with the MPC-based controllers. However, as the complexity of the environment increases to 3) and 4), the PD and AC controllers are no longer stable, and the tracking accuracy of MPC also becomes worse, while the RMMPC can still maintain a small tracking error on the order of millimeter since it can model the uncertainty in the environment. The results show that the proposed MMC-based controller design can improve the robustness of the manipulation in the presence of varying environmental resistance.

%

\subsection{Real-world Experiments}

\begin{figure*}[t]
\setlength{\abovecaptionskip}{-0.1cm}
\centering
\includegraphics[scale=1.0,angle=0,width=0.98\textwidth]{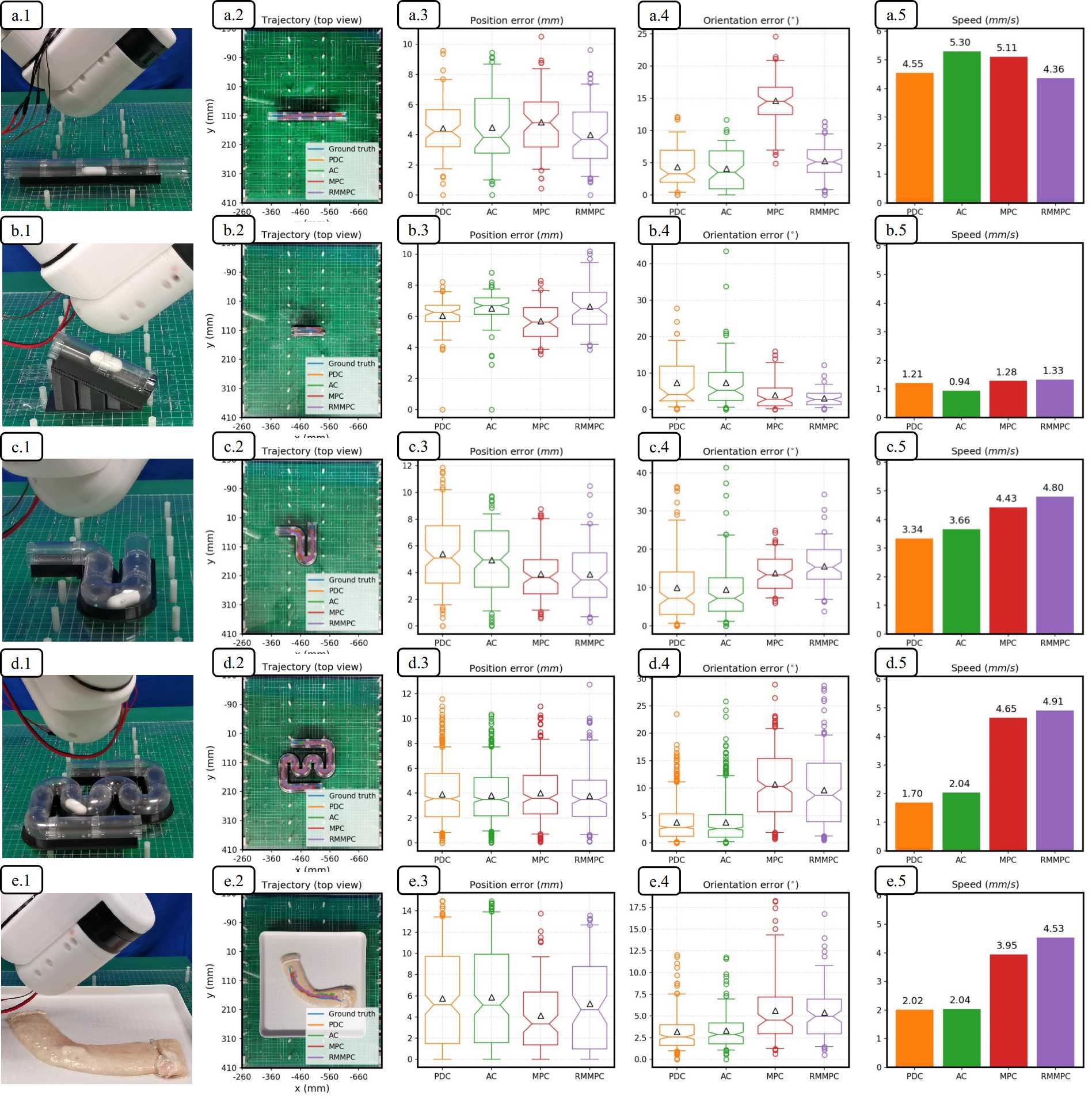}
\caption{The real-world experiments are conducted in (a) a straight PVC tube, (b) a straight PVC tube placed over a $30^{\circ}$ tubular slope, (c) a PVC tube with a $90^{\circ}$ bend and a $180^{\circ}$ sharp bend, (d) a more complex-shaped PVC tube with several sharp bends, and (e) a curved ex-vivo pig colon. The top view of the tracking trajectories, the position accuracy, orientation accuracy and time consumption during trajectory following in the five environments are shown in the second, third, fourth and fifth column, respectively.}
\label{Fig_expdemo}
\vspace{-0.2cm}
\end{figure*}

In order to further evaluate the performance of the four trajectory following approaches, we conduct $5$ sets of real-world experiments in $5$ different tubular environments with increasing complexity, as shown in Fig. \ref{Fig_expdemo}. Each row corresponds to one set of experiments conducted in one tubular environment ((a)-(d) are phantoms with different shapes and (e) is an ex-vivo pig colon). The first column shows the snapshots of each set of experiments. The second column the top view of the experiments and the estimated trajectories of the capsule under different control strategies. The trajectory tracking errors in position and orientation are presented in the third and fourth columns, respectively, and the actual moving speed of the capsule under different control methods is compared in the last column. Before each set of experiments, the GT trajectory is manually specified according to the centerline of each tubular environment. The pre-set constant speed of the capsule is set to $5mm/s$.

As shown in Fig. \ref{Fig_expdemo}(a), the capsule is actuated to move through a straight PVC tube with a length of $215mm$ following a pre-defined trajectory. All the four controllers can effectively track the desired trajectory with a similar average position error of less than $5mm$, and the PD, AC and RMMPC can achieve a small average orientation error of less than $5^{\circ}$. In Fig \ref{Fig_expdemo} (a.5), the moving speed of the capsule is close to the pre-set value (around $5mm/s$) with the four control methods.

Fig. \ref{Fig_expdemo}(b) presents the experimental results in a straight PVC tube (length $50mm$) placed over a $30^{\circ}$ slope. All the controllers can complete the trajectory following task with a similar average position error of less than $7mm$ and orientation error of less than $10^{\circ}$. However, it can be observed in Fig. \ref{Fig_expdemo} (b.5) that the moving speed of the capsule is relatively slow (around $1mm/s$) for all the control methods, which is mainly because the gravity of the capsule is not perfectly compensated by the magnetic force during rotation due to the approximation error in (\ref{F_f_theta_c_approximate}). Nevertheless, the results in this experiments preliminarily demonstrate that our approach can successfully complete the trajectory following task in a tubular environment with varying altitudes.

As shown in Fig. \ref{Fig_expdemo}(c), the capsule is actuated in a PVC lumen with a $180^{\circ}$ bend and a $90^{\circ}$ bend (total length $244mm$). As shown in Fig. \ref{Fig_expdemo} (c.3) and (c.4), the MPC and RMMPC methods achieve better accuracy in position than the reactive control methods, but their tracking errors in orientation are slightly larger than the PD and AC. It can be found that the RMMPC has the best performance in terms of time consumption among all the controllers.

In Fig. \ref{Fig_expdemo}(d), when the capsule is moving through a more complex-shaped PVC lumen with several $180^{\circ}$ sharp bends (length $684mm$), all controllers show similar performance in position accuracy, and the reactive controllers have slightly better orientation accuracy. However, as can be observed in Fig. \ref{Fig_expdemo} (d.5), the MPC and RMMPC can achieve much higher actuation speed of the capsule (close to the pre-set speed $5mm/s$) than the PD and AC, since they can anticipate the future states of the system based on the dynamic model.

Finally, we actuate the capsule to follow a pre-defined trajectory under RRMA in an ex-vivo pig colon, which is more complicated than the PVC phantoms and closer to the human intestinal environment, as shown in Fig. \ref{Fig_expdemo}(e). All the controllers have an orientation accuracy of less than $7.5^{\circ}$, and the MPC and RMMPC can achieve a better position accuracy than the PD and AC. Besides, as shown in Fig. \ref{Fig_expdemo} (e.5), the MPC and RMMPC have much better performance in time consumption compared with PD and AC controllers.

From Fig. \ref{Fig_expdemo} (a)(c)(d)(e), we can observe that as the complexity of the environment increases, the anticipative controllers (i.e., MPC and RMMPC) can generally achieve better performance than the reactive controllers (i.e., PD and AC), especially in the time consumption. This is because these controllers not only utilize the error between the current pose and the desired pose of the capsule for control, but also take the future trajectory into consideration. Since the RMMPC can take into account the uncertainty of the environmental resistance, it can better deal with the varying shapes and materials of the tubular environments and maintain a tracking speed of $5mm/s$ across different environments, and also achieves a high tracking accuracy of $5.26\pm4.32mm$ and $5.38^{\circ}\pm3.16^{\circ}$ in the ex-vivo pig colon. These results have demonstrated the effectiveness of our proposed approach in the challenging task of manipulating a capsule in complex tubular environments with intestine-like shapes and materials. 

Assuming that the moving speed of the capsule in the human small intestine is the same as that in the pig colon, it can be roughly estimated that it would take about $22$ minutes to traverse the small intestine (with a length of about $6m$ \cite{hounnou2002anatomical}) with active WCE using our proposed system.
Compared with the long duration of current passive WCE (which takes $\sim 247.2$ min to examine the small intestine \cite{liao2010fields}), the proposed technology holds great promise to significantly shorten the examination time, enable efficient and accurate inspection of the intestine, and improve the clinical acceptance of this non-invasive and painless screening technique.

\begin{table*}[t] \scriptsize \renewcommand\arraystretch{1.1}
\centering 
\caption{Comparison of Permanent Magnet-based Systems for Trajectory Following of A Capsule}
\resizebox{\textwidth}{40mm}{
\begin{tabular}{|m{1cm}<{\centering}|m{1.1cm}<{\centering}|m{1.8cm}<{\centering}|m{1cm}<{\centering}|m{1.8cm}<{\centering}|m{5.5cm}<{\centering}|m{5cm}<{\centering}|}
\hline
{\textbf{Study}} & {\textbf{Actuation method}} & {\textbf{Localization feedback}} & {\textbf{Controller}} & {\textbf{Experimental environment}}& {\textbf{Task description}} & {\textbf{Tracking accuracy (position / orientation)}} \\
\hline
\hline
{Mahoney et al. \cite{mahoney2016five}} & Dragging magnetic actuation & External visual sensors & {PID} & In a water-filled tank & Follow a pre-set trajectory on the y-z plane & $2.8mm$ / Not reported \\
\hline
{Taddese et al. \cite{taddese2018enhanced}} & Dragging magnetic actuation & Internal magnetic sensors and IMU & PID & {Between two acrylic planes} & {Follow a pre-set straight trajectory} & {$-5.30\pm2.6 mm$ / $4.96\pm2.2^\circ$} \\
\hline
{Pittiglio et al. \cite{pittiglio2019magnetic}} & Dragging magnetic actuation & Internal magnetic sensors and IMU & {PD} & In a colon phantom  & Follow a pre-set trajectory with gravity compensation & Localization accuracy: $4.0 mm$ / Not reported; Tracking accuracy is not reported \\
\hline
{Barducci et al. \cite{barducci2019adaptive}} & Dragging magnetic actuation & Internal magnetic sensors and IMU & {AC} & In a colon phantom  &Follow a pre-set trajectory with gravity compensation& Localization accuracy: $4.0 mm$ / Not reported; Tracking accuracy is not reported\\
\hline
\multirow{3}{*}{\tabincell{c}{Scaglioni \\ et al. \cite{scaglioni2019explicit}}} &  \multirow{3}{*}{\tabincell{c}{Dragging \\magnetic\\ actuation}} & \multirow{3}{*}{\tabincell{c}{Internal magnetic \\sensors and IMU}} & \multirow{3}{*}{\tabincell{c}{{MPC}}} & \multirow{3}{*}{\tabincell{c}{In a colon \\phantom}}  & Follow a pre-set trajectory through a straight lumen with an obstacle & {$35.0\pm4.8 mm$ / $4.8\pm0.6^\circ$} \\
\cline{6-7}
{} & {} &{} &{} &{} &Follow a desired trajectory through a curved lumen with a $90^\circ$ bend &{$11.0\pm5.8 mm$ / $5.2\pm0.7 ^\circ$} \\
\hline
Norton et al. \cite{norton2019intelligent} & Dragging magnetic actuation & Internal magnetic sensors and IMU & {PD} & {In an in-vivo porcine model} & Follow a pre-set linear trajectory for ultrasound acquisitions &  Localization accuracy: $2.0 mm$ / $3.0^\circ$; Tracking accuracy is not reported \\
\hline
\multirow{10}{*}{\tabincell{c}{Ours}} & \multirow{8}{*}{\tabincell{c}{Reciprocally \\rotating \\magnetic\\ actuation}} & \multirow{8}{*}{\tabincell{c}{External magnetic \\ sensors}}  & \multirow{8}{*}{\tabincell{c}{RMMPC}} & \multirow{6}{*}{\tabincell{c}{In PVC phantoms}} & {Follow a preset trajectory through a straight lumen} &  $4.00\pm2.02mm$ / $5.29\pm2.56^\circ$  \\
\cline{6-7}
{} & {} & {} & {}&{}&Follow a preset trajectory to climb a $30^\circ$ slope  & $6.65\pm1.64mm$ / $3.23\pm2.38^\circ$\\
\cline{6-7}
{} & {} & {}  & {}&{}&Follow a preset trajectory through a curved lumen with a $90^\circ$ bend and a $180^\circ$ bend  &$3.88\pm2.32mm$ / $15.60\pm5.85^\circ$\\
\cline{6-7}
{} & {} & {}  & {}&{}&Follow a preset trajectory through a complex-shaped lumen with several sharp bends  &$3.79\pm2.29mm$ / $9.67\pm6.33^\circ$ \\
\cline{5-7}
{} & {} & {} & {} & In an ex-vivo pig colon & Follow a preset trajectory through a curved pig colon&{$5.26\pm4.32mm$ / $5.38\pm3.16^\circ$} \\
\hline
\end{tabular}}
\label{T_comparison}
\end{table*}

A detailed comparison between our method and previous studies is listed in Table \ref{T_comparison}. It can be seen that our method achieves the state-of-the-art tracking accuracy in tubular (or colon-like) environments, using the proposed RRMA-based trajectory following strategies. Also, since our method only uses external magnetic sensors for capsule localization, the size and power consumption of the capsule can be reduced compared with the internal sensor-based methods.

\subsection{Video Demonstration}

The video demo can be found at \url{https://youtu.be/Xa9XtI4KToY} for a better visualization of our results.

\section{Conclusions}

In this paper, we have investigated the trajectory following problem of a robotic capsule endoscope under RRMA in a tubular environment based on four different controllers, and implemented the methods on an external sensor-based SMAL system. Our methods are validated in simulation and real-world experiments in several complex-shaped tubular environments. The results show that our proposed trajectory following strategies for active WCE can effectively and robustly actuate a capsule to follow a pre-set trajectory in a complex-shaped tubular environment, which has the potential to realize fast and accurate inspection of the intestine at given points, thereby improving the diagnostic accuracy and efficiency of WCE. Although the work presented in this paper is targeted at the magnetic manipulation of capsule endoscopes in the intestinal environment, the methods can also be applied to the magnetic control of medical robots working in tubular environments in general.
In the future, the proposed approach needs to be further evaluated and improved based on in-vivo trials to towards a potential clinical translation.

\bibliographystyle{IEEEtran}
\bibliography{root}

\end{document}